\documentclass[letterpaper, 10 pt, conference]{ieeeconf}

\IEEEoverridecommandlockouts
\usepackage{amsmath}
\usepackage{amssymb}
\usepackage{graphicx}

\title{\LARGE \bf
Unified Walking, Running, and Recovery for Humanoids via State-Dependent Adversarial Motion Priors
}

\author{Yidan Lu$^{}$, Yichao Zhong$^{}$, Liu Zhao, Wanyue Li, Peng Lu$^{\dagger}$%
\thanks{\raggedright $^{\dagger}$Corresponding author: Peng Lu ({\tt\small lupeng@hku.hk}).}%
\thanks{\raggedright The University of Hong Kong, Hong Kong SAR, China.\protect\newline
        \textit{E-mail:}\ {\tt\small ydlu@connect.hku.hk}, {\tt\small yichao.zhong@connect.hku.hk}, {\tt\small zhaol@connect.hku.hk},\protect\newline
        {\tt\small liwy1024@connect.hku.hk}.}%
}

\begin{document}

\maketitle
\thispagestyle{empty}
\pagestyle{empty}

\begin{abstract}
We propose a unified reinforcement learning framework that enables a single policy to perform walking, running, and fall recovery on the Unitree G1 humanoid robot, validated on physical hardware without any explicit mode-switching command at deployment. The framework extends Adversarial Motion Priors (AMP) by replacing the conventional global reference distribution with a state-dependent gate that routes each training transition to one of two discriminators: a dedicated recovery discriminator and a velocity-conditioned locomotion discriminator that jointly covers walking and running. The gate is defined by a single fixed threshold on projected gravity: the recovery discriminator is activated when body tilt exceeds approximately $37^\circ$ from vertical ($|g_z+1|>0.6$); otherwise the locomotion discriminator is used, with the normalized commanded velocity serving as a condition that selects the appropriate reference trajectory between walk and run clips. Only three LAFAN1 reference clips are required to regularize the complete behavior set. At deployment, a single frozen ONNX policy executes at 50\,Hz with no runtime mode logic; hardware experiments demonstrate successful recovery from both prone and supine falls and smooth walk-to-run transitions under the same controller.
\end{abstract}

\section{INTRODUCTION}

Humanoid robots deployed in unstructured environments must seamlessly execute agile locomotion and robustly recover from falls. Conventional approaches address these demands through modular architectures in which walking, running, and recovery are implemented as separate controllers coordinated by a hand-crafted finite state machine~\cite{capturepoint}. While this decomposition simplifies individual controller design, it entails significant practical limitations: transition logic requires manual tuning, behavioral boundaries can be brittle near mode boundaries, and motion quality may degrade during inter-module handoffs.

Recent advances in reinforcement learning have demonstrated that humanoid standing-up and fall recovery can be acquired end-to-end without handcrafted motion scripts~\cite{host,humanup,hifar,frasa,safefall,unifiedfallsafety}. In parallel, motion-prior-based methods such as DeepMimic and Adversarial Motion Priors (AMP) have established that reference motions significantly improve the physical plausibility and style consistency of learned behaviors~\cite{deepmimic,amp}. However, AMP is most naturally formulated for a single behavioral mode or a shared motion distribution, whereas humanoid operation in practice spans qualitatively distinct regimes—steady locomotion, high-speed running, and recovery from failure—each with different kinematic characteristics.

This work proposes a targeted extension: maintaining a single shared actor policy while switching the adversarial motion prior in response to the current robot state and command. Two discriminators are trained in parallel—a recovery discriminator regularized by a fall-recovery reference clip, and a velocity-conditioned locomotion discriminator that jointly covers walking and running by conditioning on the commanded velocity to select the appropriate reference trajectory. A lightweight, interpretable gating function selects the active discriminator at each training step based on body orientation, ensuring that the AMP reward signal is always mode-consistent. The resulting policy requires no runtime mode selector and is deployed as a single frozen model, yet produces behaviorally appropriate motion across all three regimes.

\begin{figure}[t]
    \centering
    \includegraphics[width=\columnwidth]{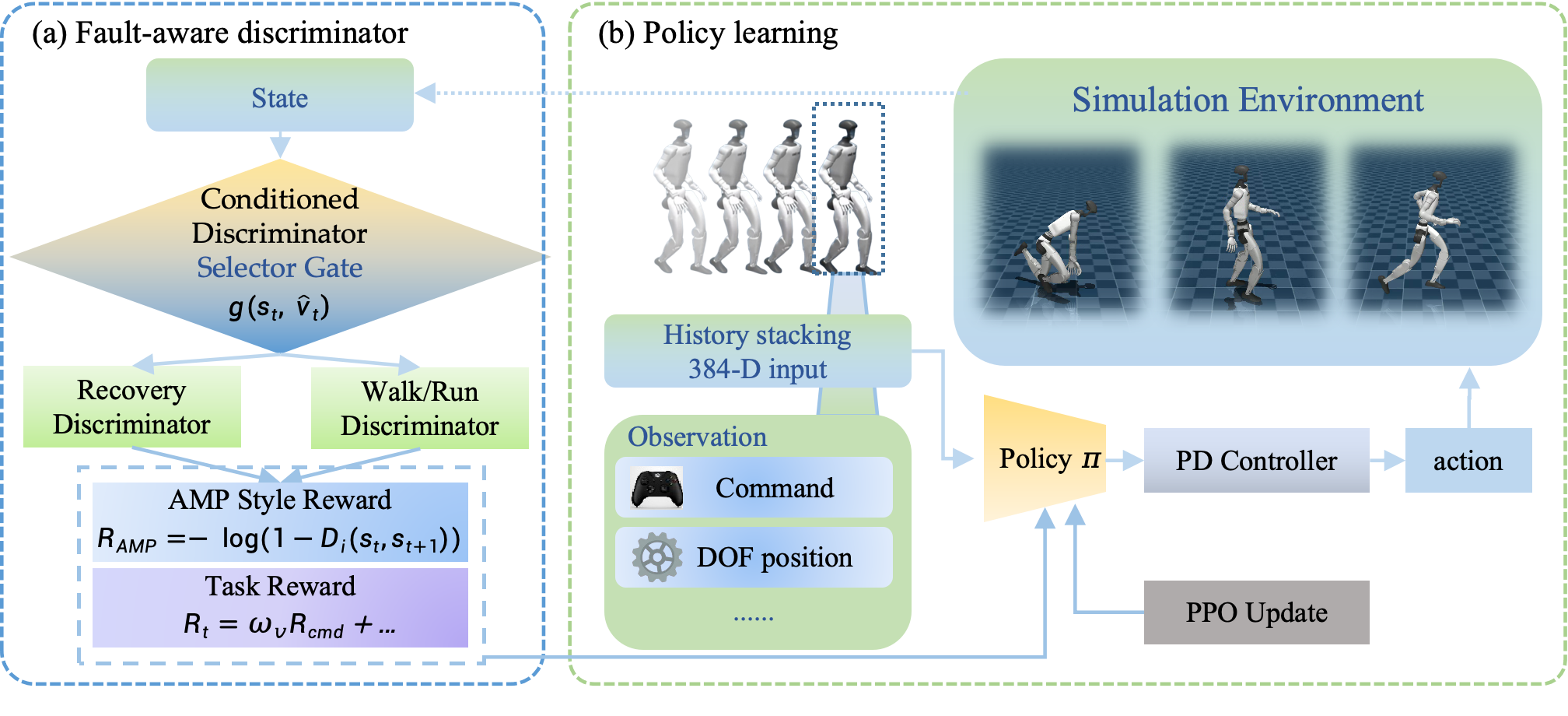}
    \caption{Overview of the proposed framework. A shared actor policy $\pi_\theta$ is trained with a composite reward comprising task terms and a mode-specific AMP reward. At each training step, the state-dependent gate $g(s_t, \hat{v}_t)$ routes each transition to one of two discriminators: the recovery discriminator (activated when body tilt exceeds ${\sim}37^\circ$) or the velocity-conditioned locomotion discriminator, which selects between walk and run reference trajectories based on the normalized commanded velocity $\hat{v}_t$. At deployment, only the frozen policy is used; no runtime mode logic is required.}
    \label{fig:framework}
\end{figure}

\begin{figure*}[t]
    \centering
    \includegraphics[width=\textwidth]{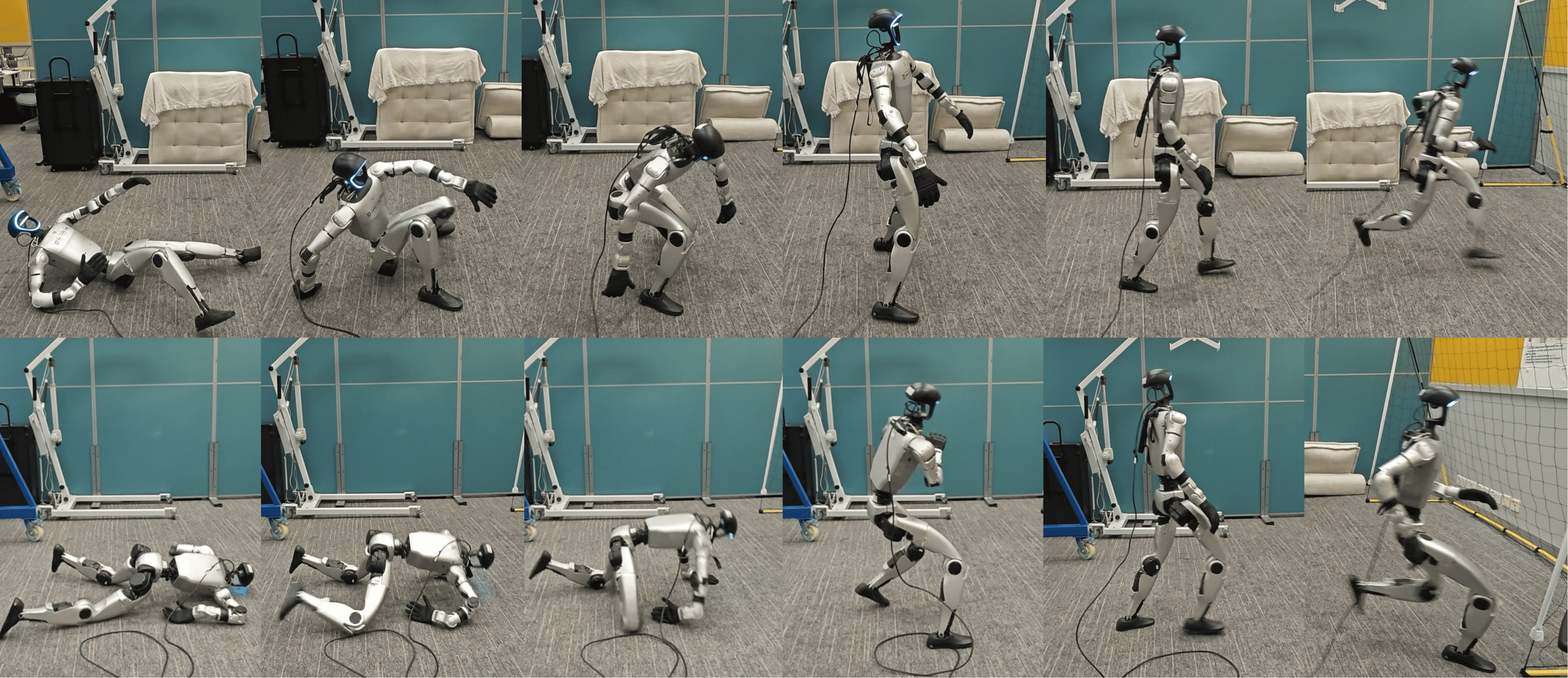}
    \caption{Hardware demonstration of the unified policy on Unitree G1. \textbf{Top:} supine recovery$\rightarrow$walk$\rightarrow$run. \textbf{Bottom:} prone recovery$\rightarrow$walk$\rightarrow$run. Both sequences are executed by a single deployed controller with no mode-switch command; the active motion prior is selected automatically by the state-dependent gate (Eq.~\ref{eq:gate}).}
    \label{fig:teaser}
\end{figure*}

\section{PROBLEM FORMULATION}

We consider the problem of learning a single humanoid controller $\pi_\theta(a_t \mid o_t)$ that spans three qualitatively distinct behavioral regimes: walking, running, and fall recovery. The observation $o_t$ comprises body angular velocity, projected gravity, commanded velocity, relative joint positions, joint velocities, and previous actions. Each observation frame is 96-dimensional; a history of four consecutive frames is concatenated before policy inference, yielding a 384-dimensional input. The action $a_t \in \mathbb{R}^{29}$ specifies target joint positions, which are tracked by low-level PD controllers.

The central challenge is maintaining coherent behavioral regularization across regimes with fundamentally different kinematics: walking and running are cyclic velocity-tracking behaviors, whereas recovery requires rapid posture reorientation through multi-contact support phases. A single shared motion prior is poorly suited to this envelope—a global reference distribution either fails to capture high-speed running statistics or conflates locomotion with recovery dynamics, degrading motion quality in both regimes.

\section{METHOD}

\subsection{Policy and Task Reward}

The actor policy $\pi_\theta(a_t \mid o_t)$ is trained end-to-end and deployed as a single ONNX model without any runtime mode logic. The primary task reward combines velocity tracking, motion smoothness, postural stability, and safety penalties:
\begin{equation}
\begin{aligned}
R_t ={}& w_v R_{\text{cmd}} + w_s R_{\text{smooth}} + w_p R_{\text{posture}} \\
&- w_e C_{\text{energy}} - w_f C_{\text{fall}}.
\end{aligned}
\end{equation}

\subsection{State-Dependent Motion Prior}

We employ two discriminators. A recovery discriminator $D_\phi^{\text{rec}}$ is trained exclusively on fall-recovery transitions. A velocity-conditioned locomotion discriminator $D_\phi^{\text{loco}}$ accepts the normalized commanded velocity $\hat{v}_t \in [0,1]$ as an additional input and is trained jointly on walk and run reference transitions, using $\hat{v}_t$ to interpolate between the two reference distributions—providing style-consistent rewards across the full speed range without separate walk and run networks.

For each state transition $(s_t, s_{t+1})$ encountered during training, a binary gating function $g(s_t)$ selects the active discriminator:
\begin{equation}
z_t = g(s_t) \in \{\text{rec},\; \text{loco}\}.
\end{equation}
The AMP style reward is then computed as
\begin{equation}
R_{\text{AMP}} =
\begin{cases}
-\log\!\left(1 - D_\phi^{\text{rec}}(s_t, s_{t+1})\right) & \text{if } z_t = \text{rec}, \\[4pt]
-\log\!\left(1 - D_\phi^{\text{loco}}(s_t, s_{t+1} \mid \hat{v}_t)\right) & \text{if } z_t = \text{loco},
\end{cases}
\end{equation}
and the total training reward is
\begin{equation}
R_t^{\text{total}} = R_t + \lambda_{\text{amp}}\, R_{\text{AMP}}.
\end{equation}

\subsection{Gating Function}

The gate $g$ is defined by a single fixed threshold rather than a learned classifier:
\begin{equation}
z_t =
\begin{cases}
\text{rec}  & \text{if } |g_z + 1| > 0.6, \\
\text{loco} & \text{otherwise,}
\end{cases}
\label{eq:gate}
\end{equation}
where $g_z$ is the $z$-component of the projected gravity vector ($g_z \approx -1$ when upright, near zero or positive when fallen). A fixed threshold is justified because the gravity signal is bimodal by construction—the $|g_z+1|>0.6$ boundary corresponds to ${\approx}37^\circ$ tilt and lies in a low-occupancy region of the signal's empirical distribution, making it accurate and robust to noise without a learned gating mechanism. When the locomotion branch is active, $\hat{v}_t = \min(v_x^{\mathrm{cmd}} / v_{\max}, 1)$ is passed as a condition to $D_\phi^{\text{loco}}$, selecting reference transitions from the walk clip at low $\hat{v}_t$ and from the run clip at high $\hat{v}_t$. This gate is active only during training; at inference time, the deployed policy is a monolithic network with no explicit mode variable.

\subsection{Reference Motion Retargeting}

Three reference clips are drawn from the LAFAN1 dataset~\cite{lafan1} and retargeted to the G1 joint space via inverse kinematics: \texttt{walk1\_subject1}, \texttt{run1\_subject2}, and \texttt{fallAndGetUp2\_subject2}. The recovery clip exclusively trains $D_\phi^{\text{rec}}$. Both locomotion clips train $D_\phi^{\text{loco}}$: at each discriminator update, a reference transition is sampled from the walk clip with probability $(1{-}\hat{v}_t)$ and from the run clip with probability $\hat{v}_t$, so the condition consistently indexes the intended reference distribution.

\section{EXPERIMENTS}

\subsection{Setup}
The policy is trained in Isaac Lab~\cite{isaaclab} simulation using Proximal Policy Optimization (PPO)~\cite{ppo} with AMP auxiliary reward weighted at $\lambda_{\mathrm{amp}} = 0.5$. The network is a three-layer MLP operating on a 384-dimensional stacked observation. Upon convergence, the policy is exported as an ONNX model and deployed at 50\,Hz on the physical Unitree G1 via a C++ finite state machine that reads joint states and issues PD position targets. No sim-to-real adaptation or domain randomization fine-tuning is applied beyond standard simulation parameter randomization during training.

\subsection{Locomotion}
The deployed policy successfully tracks forward velocity commands over $[-0.5, 1.0]$\,m/s in normal operating mode and $[-1.5, 3.0]$\,m/s in fast mode. Fast mode is engaged via an explicit operator command, reflecting a deliberate safety constraint. Across this speed range, the velocity-conditioned locomotion discriminator $D_\phi^{\text{loco}}$ continuously interpolates between walk and run reference distributions via $\hat{v}_t$, producing a smooth style-consistent reward signal without any discrete mode switch. Observed gait characteristics are consistent with the respective reference clips: walking exhibits a regular heel-toe contact cycle, while running displays elongated stride length and reduced double-support phase duration.

\subsection{Fall Recovery}
Recovery is evaluated from two hardware configurations: prone (face-down) and supine (face-up). Because the projected gravity condition in Eq.~\eqref{eq:gate} continuously routes fallen-state transitions to $D_\phi^{\text{rec}}$ throughout training, the policy acquires recovery behavior without any dedicated recovery module or separate training phase. As shown in Fig.~\ref{fig:teaser}, the robot successfully regains a stable upright stance from both configurations without manual intervention.

\subsection{Unified Behavioral Continuity}
Fig.~\ref{fig:teaser} presents complete hardware rollouts demonstrating the unified nature of the controller. Starting from a fallen configuration, the robot executes the full recovery$\rightarrow$walk$\rightarrow$run sequence under a single deployed policy in both the supine and prone conditions. These results confirm that the three behavioral regimes are stably encoded within a single policy and that the mode transitions induced by the training-time gating rule produce coherent, continuous motion at deployment.

\section{CONCLUSION}

We have presented a unified humanoid control framework in which a single reinforcement learning policy covers walking, running, and fall recovery through state-dependent adversarial motion priors. The key design choice is a two-discriminator structure: a dedicated recovery discriminator and a velocity-conditioned locomotion discriminator that jointly handles walking and running within a single network. A single fixed threshold on projected gravity routes training transitions between the two discriminators, ensuring mode-consistent style regularization without learned gating complexity. The minimality of the reference set—three LAFAN1 clips—demonstrates that structural separation of recovery and locomotion priors can be more important than reference diversity for achieving qualitatively distinct and physically plausible behaviors within a shared policy. Hardware validation on the Unitree G1 confirms that the learned controller generalizes from simulation to real-world conditions and supports seamless behavioral transitions across locomotion and recovery regimes without runtime mode management.

\bibliographystyle{IEEEtran}
\bibliography{reference}

\end{document}